\documentclass[11pt,a4paper]{article}
\usepackage[hyperref]{acl2021}
\usepackage{times}
\usepackage{latexsym}

\usepackage{microtype}
\usepackage{graphicx}
\usepackage{amsmath,mathtools}
\usepackage{amssymb}
\usepackage{xcolor}
\usepackage{algorithm, algorithmic}
\usepackage{multirow}
\usepackage{textcomp,booktabs}
\usepackage{color}
\usepackage{colortbl}
\definecolor{mygray}{gray}{.9}

\aclfinalcopy 

\title{Directed Acyclic Graph Network for Conversational Emotion Recognition}

\author{Weizhou Shen,  Siyue Wu,  Yunyi Yang, Xiaojun Quan\thanks{\; Corresponding author.} \\
  School of Computer Science and Engineering, Sun Yat-sen University, China \\
  \texttt{ \{shenwzh3, wusy39, yangyy37\}@mail2.sysu.edu.cn}\\
  \texttt{quanxj3@mail.sysu.edu.cn}}

\date{}

\begin{document}
\maketitle
\begin{abstract}
The modeling of conversational context plays a vital role in emotion recognition from conversation (ERC). In this paper, we put forward a novel idea of encoding the utterances with a directed acyclic graph (DAG) to better model the intrinsic structure within a conversation, and design a directed acyclic neural network,~namely DAG-ERC\footnote{The code is available at \url{https://github.com/shenwzh3/DAG-ERC}}, to implement this idea.~In an attempt to combine the strengths of conventional graph-based neural models and recurrence-based neural models,~DAG-ERC provides a more intuitive way to model the information flow between long-distance conversation background and nearby context.~Extensive experiments are conducted on four ERC benchmarks with state-of-the-art models employed as baselines for comparison.~The empirical results demonstrate the superiority of this new model and confirm the motivation of the directed acyclic graph architecture for ERC.

\end{abstract}

\section{Introduction}
Utterance-level emotion recognition in conversation (ERC) is an emerging task that aims to identify the emotion of each utterance in a conversation. This task has been recently concerned by a considerable number of NLP researchers due to its potential applications in several areas, such as opinion mining in social media \citep{chatterjee2019semeval} and building an emotional and empathetic dialog system \citep{majumder2020mime}. 

The emotion of a query utterance is likely to be influenced by many factors such as the utterances spoken by the same speaker and the surrounding conversation context. Indeed, how to model the conversational context lies at the heart of this task \citep{poria2019emotion}. Empirical evidence also shows that a good representation of conversation context significantly contributes to the model performance, especially when the content of query utterance is too short to be identified alone \citep{ghosal2019dialoguegcn}.

Numerous efforts have been devoted to the modeling of conversation context. Basically, they can be divided into two categories: graph-based methods \citep{zhang2019modeling, ghosal2019dialoguegcn,zhong2019knowledge,ishiwatari2020relation,shen2020dialogxl} and recurrence-based methods \citep{hazarika2018icon, hazarika2018conversational, majumder2019dialoguernn, ghosal2020cosmic}. 
For the graph-based methods, they concurrently gather information of the surrounding utterances within a certain window, while neglecting the distant utterances and the sequential information. For the recurrence-based methods, they consider the distant utterances and sequential information by encoding the utterances temporally. However, they tend to update the query utterance's state with only relatively limited information from the nearest utterances, making them difficult to get a satisfying performance.


\begin{figure}[t]
	\centering
	\includegraphics[scale=0.13]{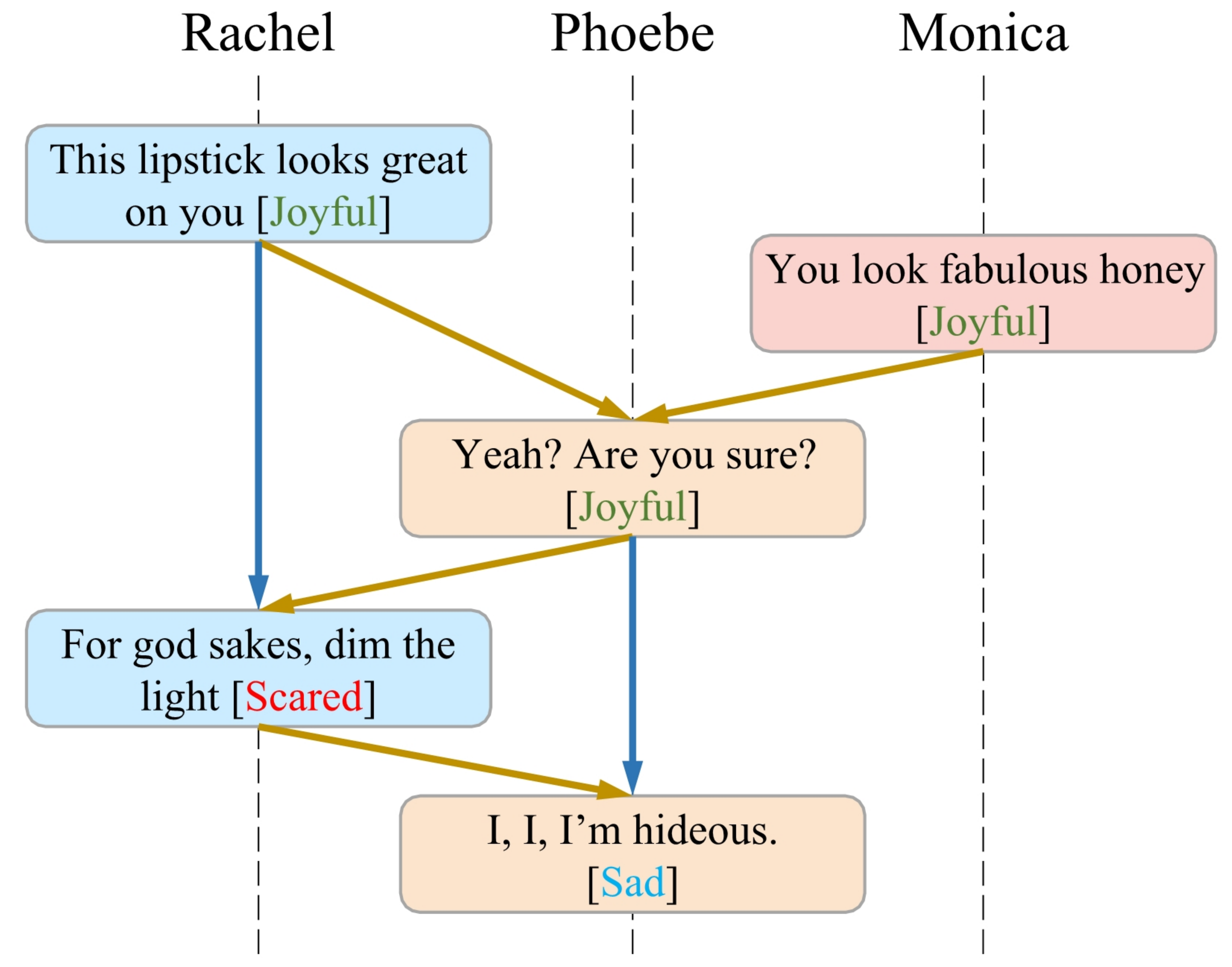} 
	\caption{Conversation as a directed acyclic graph, with brown directed edges representing  the information propagation between speakers and blue ones representing the information propagation inside a same speaker.}
	\label{fig:introduction}
	\vspace{-0.4cm}
\end{figure} 

According to the above analysis, an intuitively better way to solve ERC is to allow the advantages of graph-based methods and recurrence-based models to complement each other. This can be achieved by regarding each conversation as a directed acyclic graph (DAG).
As illustrated in Figure \ref{fig:introduction}, each utterance in a conversation only receives information from some previous utterances and cannot propagate information backward to itself and its predecessors through any path. 
This characteristic indicates that a conversation can be regarded as a DAG. 
Moreover, by the information flow from predecessors to successors through edges, DAG can gather information for a query utterance from both the neighboring utterances and the remote utterances, which acts like a combination of graph structure and recurrence structure. Thus, we speculate that DAG is a more appropriate and reasonable way than graph-based structure and recurrence-based structure to model the conversation context in ERC.

In this paper, we propose a method to model the conversation context in the form of DAG. Firstly, rather than simply connecting each utterance with a fixed number of its surrounding utterances to build a graph, we propose a new way to build a DAG from the conversation with constraints on speaker identity and positional relations. 
Secondly, inspired by DAGNN \citep{thost2021directed}, we propose a directed acyclic graph neural network for ERC, namely DAG-ERC. Unlike the traditional graph neural networks such as GCN \citep{kipf2016semi} and GAT \citep{velivckovic2017graph} that aggregate information from the previous layer, DAG-ERC can recurrently gather information of predecessors for every utterance in a single layer, which enables the model to encode the remote context without having to stack too many layers. Besides, in order to be more applicable to the ERC task, our DAG-ERC has two improvements over DAGNN: (1) a relation-aware feature transformation to gather information based on speaker identity and (2) a contextual information unit to enhance the information of historical context. We conduct extensive experiments on four ERC benchmarks and the results show that the proposed DAG-ERC achieves comparable performance with the state-of-the-art models. Furthermore, several studies are conducted to explore the effect of the proposed DAG structure and the modules of DAG-ERC.

The contributions of this paper are threefold. First, we are the first to consider a conversation as a directed acyclic graph in the ERC task. Second, we propose a method to build a DAG from a conversation with constraints based on the speaker identity and positional relations. Third, we propose a directed acyclic graph neural network for ERC, which takes DAGNN as its backbone and has two main improvements designed specifically for ERC.

\section{Related work}

\subsection{Emotion Recognition in Conversation} 

Recently, several ERC datasets with textual data have been released \citep{busso2008iemocap, schuller2012avec,zahiri2017emotion, li2017dailydialog, chen2018emotionlines, poria2019meld}, arousing the widespread interest of NLP researchers. In the following paragraphs, we divide the related works into two categories according to the methods they use to model the conversation context.

\noindent \textbf{Graph-based Models}
DialogGCN \citep{ghosal2019dialoguegcn} treats each dialog as a graph in which each utterance is connected with the surrounding utterances. RGAT \citep{ishiwatari2020relation} adds positional encodings to DialogGCN. ConGCN \citep{zhang2019modeling} regards both speakers and utterances as graph nodes and makes the whole ERC dataset a single graph. KET \citep{zhong2019knowledge} uses hierarchical Transformers \citep{vaswani2017attention} with external knowledge. DialogXL \citep{shen2020dialogxl} improves XLNet \cite{yang2019xlnet} with enhanced memory and dialog-aware self-attention.\footnote{We regard KET and DialogXL as graph-based models because they both adopt Transformer in which self-attention can be viewed as a fully-connected graph in some sense.}

\noindent \textbf{Recurrence-based Models} In this category, ICON \citep{hazarika2018icon} and CMN \cite{hazarika2018conversational} both utilize gated recurrent unit (GRU) and memory networks.~HiGRU \citep{jiao2019higru} contains two GRUs, one for utterance encoder and the other for conversation encoder.~DialogRNN \citep{majumder2019dialoguernn} is a recurrence-based method that models dialog dynamics with several RNNs.~COSMIC \citep{ghosal2020cosmic} is the latest model, which adopts a network structure very close to DialogRNN and adds external commonsense knowledge to improve performance.

\subsection{Directed Acyclic Graph Neural Network} 

Directed acyclic graph is a special type of graph structure that can be seen in multiple areas, for example, the parsing results of source code \cite{allamanis2018survey} and logical formulas \citep{crouse2019improving}. A number of neural networks that employ DAG architecture have been proposed, such as Tree-LSTM \citep{tai2015improved}, DAG-RNN\citep{shuai2016dag}, D-VAE \citep{zhang2019d}, and DAGNN \citep{thost2021directed}. 
DAGNN is different from the previous DAG models in the model structure.
Specifically, DAGNN allows multiple layers to be stacked, while the others have only one single layer. Besides, instead of merely carrying out naive sum or element-wise product on the predecessors' representations, DAGNN conducts information aggregation using graph attention.

\section{Methodology} 

\subsection{Problem Definition}

In ERC, a conversation is defined as a sequence of utterances $\{u_1, u_2, ..., u_N\}$, where $N$ is the number of utterances. Each utterance $u_i$ consists of $n_i$ tokens, namely $u_i = \{w_{i1}, w_{i2},...,w_{in_i}\}$. A discrete value $y_i\in \mathcal{S}$ is used to denote the emotion label of $u_i$, where $\mathcal{S}$ is the set of emotion labels. The speaker identity is denoted by a function $p(\cdot)$. For example, $p(u_i)\in \mathcal{P}$ denotes the speaker of $u_i$ and $\mathcal{P}$ is the collection of all speaker roles in an ERC dataset. The objective of this task is to predict the emotion label $y_t$ for a given query utterance $u_t$ based on dialog context $\{u_1, u_2, ..., u_{N}\}$ and the corresponding speaker identity.

\subsection{Building a DAG from a Conversation}\label{sec:build_dag}
We design a directed acyclic graph (DAG) to model the information propagation in a conversation. A DAG is denoted by $\mathcal{G} = (\mathcal{V}, \mathcal{E}, \mathcal{R})$. In this paper, the nodes in the DAG are the utterances in the conversation, i.e., $\mathcal{V} = \{u_1,u_2,...,u_N\}$, and the edge $(i,j,r_{ij}) \in \mathcal{E}$ represents the information propagated from $u_i$ to $u_j$, where $r_{ij} \in \mathcal{R}$ is the relation type of the edge. The set of relation types of edges, $\mathcal{R}=\{0,1\}$, contains two types of relation: $1$ for that the two connected utterances are spoken by the same speaker, and $0$ for otherwise.

We impose three constraints to decide when an utterance would propagate information to another, i.e., when two utterances are connected in the DAG:

\noindent
\textbf{Direction:} $\forall j>i, (j,i,r_{ji})\notin\mathcal{E}$.~A previous utterance can pass message to a future utterance, but a future utterance cannot pass message backwards.

\noindent
\textbf{Remote information:~}$\exists \tau\hspace{-0.05cm}<\hspace{-0.05cm}i, p(u_\tau)\hspace{-0.05cm}=\hspace{-0.05cm}p(u_i), (\tau,i\\,r_{\tau i})\hspace{-0.04cm}\in\hspace{-0.04cm}\mathcal{E}$ and $\forall j<\tau,~(j,i,r_{ji})\hspace{-0.04cm}\notin\hspace{-0.04cm}\mathcal{E}$.~For each utterance $u_i$ except the first one, there is a previous utterance $u_\tau$ that is spoken by the same speaker as $u_i$. The information generated before $u_\tau$ is called remote information, which is relatively less important.
We assume that when the speaker speaks $u_\tau$, she/he has been aware of the remote information before $u_\tau$.~That means, $u_\tau$ has included the remote information and it will be responsible for propagating the remote information to $u_i$.

\noindent
\textbf{Local information: }$\forall l, \tau\hspace{-0.04cm}<l\hspace{-0.04cm}<i, (l,i,r_{li})\in\mathcal{E}.$
Usually, the information of the local context is important. Consider $u_\tau$ and $u_i$ defined in the second constraint. We assume that every utterance $u_l$ in between $u_\tau$ and $u_i$ contains local information, and they will propagate the local information to $u_i$.

The first constraint ensures the conversation to be a DAG, and the second and third constraints indicate that $u_\tau$ is the cut-off point of remote and local information.~We regard $u_\tau$ as the $\omega$-th latest utterance spoken by $p(u_i)$ before $u_i$, where $\omega$ is a hyper-parameter. Then for each utterance $u_l$ in between $u_\tau$ and $u_i$, we make a directed edge from $u_l$ to $u_i$. We show the above process of building a DAG in Algorithm \ref{algo:build_DAG}.

\begin{algorithm}[t]
	\renewcommand{\algorithmicrequire}{\textbf{Input:}}
	\renewcommand{\algorithmicensure}{\textbf{Output:}}
	\caption{Building a DAG from a Conversation}
	\label{algo:build_DAG}
	\begin{algorithmic}[1]
		\REQUIRE the dialog $\{u_1, u_2, ..., u_N\}$, speaker identity $p(\cdot)$, hyper-parameter $\omega$
		\ENSURE $\mathcal{G} = (\mathcal{V},\mathcal{E},\mathcal{R})$
		\STATE $\mathcal{V} \gets \{u_1, u_2, ..., u_N\}$ 
		\STATE $\mathcal{E} \gets \emptyset$
		\STATE $\mathcal{R} \gets \{0,1\}$ 
		\FORALL{$i \in \{2,3,...,N\}$}
		\STATE $c \gets 0$
		\STATE $\tau \gets i-1$
		\WHILE{$\tau > 0$ and $c < \omega$}
		\IF{$p(u_\tau) = p(u_i)$}
		\STATE $\mathcal{E} \gets \mathcal{E}\cup\{(\tau,i,1)\}$
		\STATE $c \gets c +1$
		\ELSE
		\STATE $\mathcal{E} \gets \mathcal{E}\cup\{(\tau,i,0)\}$
		\ENDIF
		\STATE $\tau \gets \tau-1$
		\ENDWHILE
		\ENDFOR
		\STATE return $\mathcal{G} = (\mathcal{V},\mathcal{E},\mathcal{R})$
	\end{algorithmic}  
\end{algorithm}

\begin{figure}[t]
	\centering
	\includegraphics[scale=0.55]{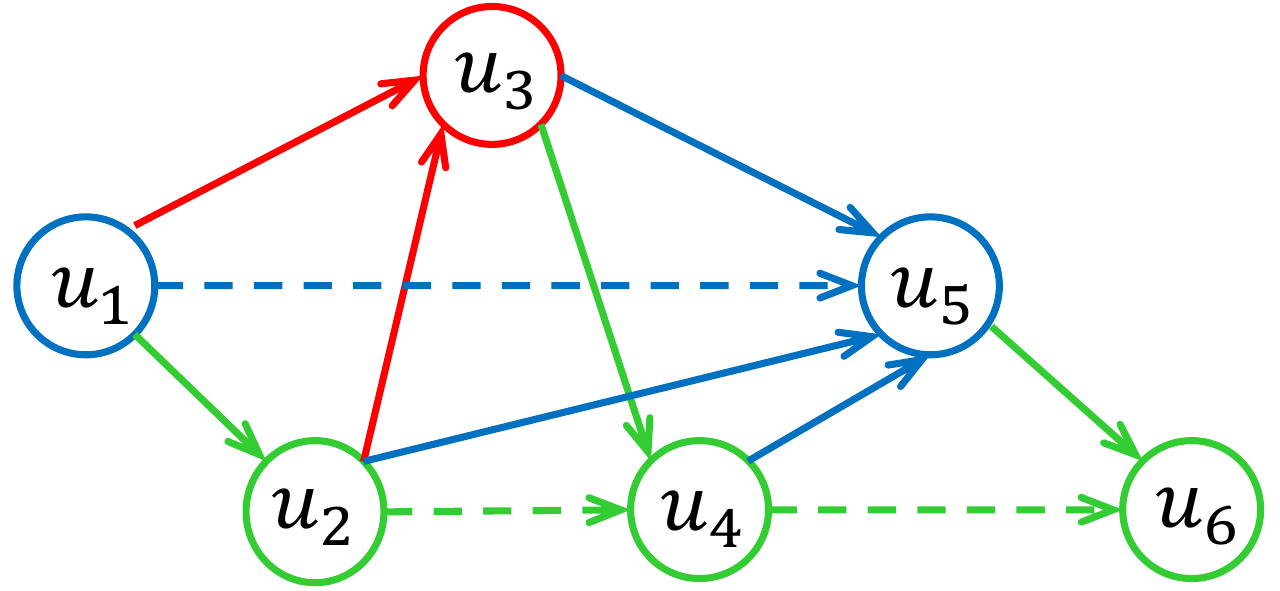}
	\caption{An example DAG built from a three-party conversation, with $\omega=1$. The three speakers' utterances are colored by red, blue and green, respectively. Solid lines represent the edges of local information, and dash lines denote the edges of remote information.}
	\label{fig:DAG}
\end{figure} 

\begin{figure*}[t]
\vspace{0.5cm}
	\centering
	\includegraphics[width=0.98\textwidth]{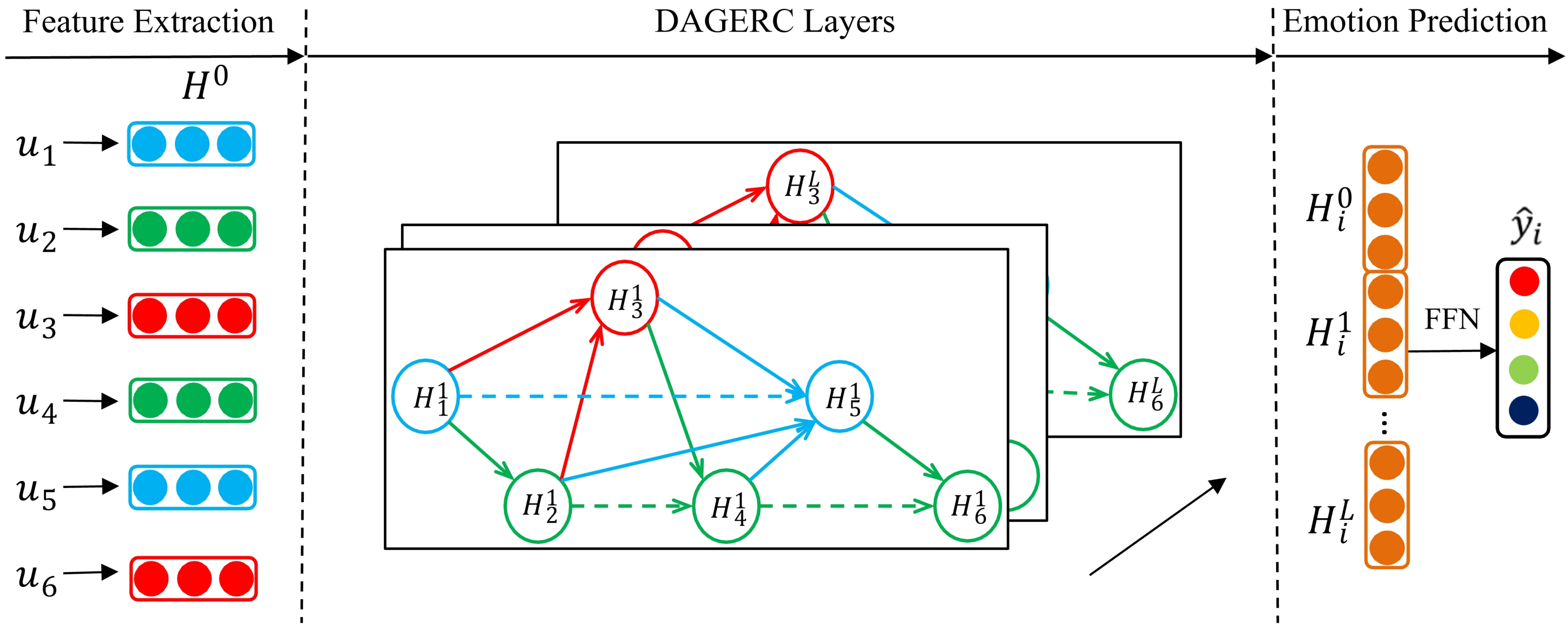} 
	\caption{The framework of {D}irected {A}cyclic {G}raph Neural Network for {ERC} (DAG-ERC).}
	\label{fig:DAG-ERC}
	\vspace{0.0cm}
\end{figure*}

An example of the DAG is shown in Figure \ref{fig:DAG}. In general, our DAG has two main advancements compared to the graph structures developed in previous works \cite{ghosal2019dialoguegcn,ishiwatari2020relation}: First, our DAG doesn't have edges from future utterances to previous utterances, which we argue is more reasonable and realistic, as the emotion of a query utterance should not be influenced by the future utterances in practice. Second, our DAG seeks a more meaningful $u_\tau$ for each utterance, rather than simply connecting each utterance with a fixed number of surrounding utterances.

\subsection{Directed Acyclic Graph Neural Network}

In this section, we introduce the proposed \textbf{D}irected \textbf{A}cyclic \textbf{G}raph Neural Network for \textbf{ERC} (DAG-ERC). The framework is shown in Figure \ref{fig:DAG-ERC}. 

\subsubsection{Utterance Feature Extraction}
DAG-ERC regards each utterance as a graph node, the feature of which can be extracted by a pre-trained Transformer-based language model. Following the convention, the pre-trained language model is firstly fine-tuned on each ERC dataset, and its parameters are then frozen while training DAG-ERC. 
Following \citet{ghosal2020cosmic}, we employ RoBERTa-Large \citep{liu2019roberta}, which has the same architecture as BERT-Large \citep{devlin2018bert}, as our feature extractor. More specifically, for each utterance $u_i$, we prepend a special token $[CLS]$ to its tokens, making the input a form of $\{[CLS],w_{i1}, w_{i2},...,w_{in_i}\}$. Then, we use the $[CLS]$'s pooled embedding at the last layer as the feature representation of $u_i$.

\subsubsection{GNN, RNN and DAGNN}\label{sec:GNN_RNN_DAGNN}

Before introducing the DAG-ERC layers in detail, we first briefly describe graph-based models, recurrence-based models and directed acyclic graph models to help understand their differences.

For each node at each layer, graph-based models (GNN) aggregate the information of its neighboring nodes at the previous layer as follows:
\begin{equation}
   \mathclap{ H^l_i = f(\text{Aggregate}(\{H^{l-1}_j|j\in\mathcal{N}_i\}),H^{l-1}_i),}\quad
\end{equation}
where $f(\cdot)$ is the information processing function, $\text{Aggregate}(\cdot)$ is the information aggregation function to gather information from neighboring nodes, and $\mathcal{N}_i$ denotes the neighbours of the $i$-th node. 

Recurrence-based models (RNN) allow information to propagate temporally at the same layer, while the $i$-th node only receives information from the $(i\hspace{-0.07cm}-\hspace{-0.07cm}1)$-th node:
\begin{equation}
    H^{l}_i = f(H^l_{i-1}, H^{l-1}_{i}).
\end{equation}

Directed acyclic graph models (DAGNN) work like a combination of GNN and RNN. They aggregate information for each node in temporal order, and allow all nodes to gather information from neighbors and update their states at the same layer:
\vspace{-0.4cm}
\begin{equation}
    H^l_i = f(\text{Aggregate}(\{H^{l}_j|j\in\mathcal{N}_i\}),H^{l-1}_i).
\end{equation}

The strength of applying DAGNN to ERC is relatively apparent: By allowing information to propagate temporally at the same layer, DAGNN can get access to distant utterances and model the information flow throughout the whole conversation, which is hardly possible for GNN. Besides, DAGNN gathers information from several neighboring utterances, which sounds more appealing than RNN as the latter only receives information from the $(i\hspace{-0.07cm}-\hspace{-0.07cm}1)$-th utterance.

\subsubsection{DAG-ERC Layers}
Our proposed DAG-ERC is primarily inspired by DAGNN \citep{thost2021directed}, with novel improvements specially made for emotion recognition in conversation. 
At each layer $l$ of DAG-ERC, due to the temporal information flow, the hidden state of utterances should be computed recurrently from the first utterance to the last one.

For each utterance $u_i$, the attention weights between $u_i$ and its predecessors are calculated by using $u_i$'s hidden state at the $(l-1)$-th layer to attend to the predecessors' hidden states at $l$-th layer:
\begin{equation}
    \alpha^l_{ij} = \text{Softmax}_{j\in\mathcal{N}_i}(W^l_\alpha[H^l_j\Vert H^{l-1}_i])
\end{equation}
where $W_\alpha^l$ are trainable parameters and $\Vert$ denotes the concatenation operation.

The information aggregation operation in DAG-ERC is different from that in DAGNN. Instead of merely gathering information according to the attention weights, inspired by R-GCN \citep{schlichtkrull2018modeling}, we apply a \emph{relation-aware feature transformation} to make full use of the relational type of edges:
\begin{equation}\label{eq:M}
    M^l_i = \sum\limits_{j\in\mathcal{N}_i}\alpha_{ij}W_{r_{ij}}^lH^l_j,
\end{equation}
where  $W_{r_{ij}}^l\in\{W_0^l,W_1^l\}$ are trainable parameters for the relation-aware transformation. 

After the aggregated information $M^l_i$ is calculated, we make it interact with $u_i$'s hidden state at the previous layer $H^{l-1}_i$ to obtain the final hidden state of $u_i$ at the current layer. 
In DAGNN, the final hidden state is obtained by allowing $M^l_i$ to control information propagation of $H^{l-1}_i$ to the $l$-th layer with a gated recurrent unit (GRU):
\begin{equation}\label{eq:DAGNN_H}
    \widetilde{H}^l_i = \text{GRU}^l_H(H^{l-1}_i, M^l_i),
\end{equation}
where $H^{l-1}_i$, $M^l_i$, and $\widetilde{H}^l_i$ are the input, hidden state and output of the GRU, respectively.

We refer to the process in Equation \ref{eq:DAGNN_H} as \emph{nodal information unit}, because it focuses on the node information propagating from the past layer to the current layer. Nodal information unit may be suitable for the tasks that DAGNN is originally designed to solve.
However, we find that only using nodal information unit is not enough for ERC, especially when the query utterance $u_i$'s emotion should be derived from its context. 
The reason is that in DAGNN, the information of context $M^l_i$ is only used to control the propagation of $u_i$'s hidden state, and under this circumstance, the information of context is not fully leveraged.~Therefore, we design another GRU called \emph{contextual information unit} to model the information flow of historical context through a single layer. In the contextual information unit, the roles of $H^{i-1}_i$ and $M^l_i$ in GRU are reversed, i.e., $H^{i-1}_i$ controls the propagation of $M^l_i$:
\begin{equation}\label{eq:info_flow}
    C^l_i = \text{GRU}^l_M(M^{l}_i, H^{l-1}_i).
\end{equation}

The representation of $u_i$ at the $l$-th layer is the sum of $\widetilde{H}^l_i$ and $C^l_i$:
\begin{equation}
    H^l_i = \widetilde{H}^l_i + C^l_i.
\end{equation}

\subsubsection{Training and Prediction}
We take the concatenation of $u_i$'s hidden states at all DAG-ERC layers as the final representation of $u_i$, and pass it through a feed-forward neural network to get the predicted emotion:
\begin{eqnarray}
&&H_i = \parallel_{l=0}^{L} H^l_i,\\
&&z_i = \text{ReLU}(W_HH_i+b_H),\\
&&P_i = \text{Softmax}(W_zz_i+b_z),\\
&&\widehat{y}_i = \text{Argmax}_{k\in\mathcal{S}}(P_i[k]).
\end{eqnarray}

For the training of DAG-ERC, we employ the standard cross-entropy loss as objective function:
\begin{equation}
\mathcal{L}(\theta) = -\sum_{i=1}^{M}\sum_{t=1}^{N_i}\text{Log}P_{i,t}[y_{i,t}],
\end{equation}
where $M$ is the number of training conversations, $N_i$ is the number of utterances in the $i$-th conversation, $y_{i,t}$ is the ground truth label, and $\theta$ is the collection of trainable parameters of DAG-ERC. 

\section{Experimental Settings}
\subsection{Implementation Details}
We conduct hyper-parameter search for our proposed DAG-ERC on each dataset by hold-out validation with a validation set. The hyper-parameters to search include learning rate, batch size, dropout rate, and the number of DAG-ERC layers.
For the $\omega$ that is described in \ref{sec:build_dag}, we let $\omega = 1$ for the overall performance comparison by default, but we report the results with $\omega$ varying from 1 to 3 in \ref{sec:variants}.
For other hyper-parameters, the sizes of all hidden vectors are equal to 300, and the feature size for the RoBERTa extractor is 1024. 
Each training and testing process is run on a single RTX 2080 Ti GPU. Each training process contains 60 epochs and it costs at most 50 seconds per epoch. The reported results of our implemented models are all based on the average score of 5 random runs on the test set.
\subsection{Datasets}
\begin{table}[t]
	\centering
	\resizebox{0.475\textwidth}{!}{
	\begin{tabular}{l|c|c|c|c|c|c}
		\toprule
		\multirow{2}*{Dataset} & \multicolumn{3}{c|}{\# Conversations} &\multicolumn{3}{c}{\# Uterrances}\\ 
		&Train&Val&Test&Train&Val&Test\\
		\hline
		IEMOCAP&\multicolumn{2}{c|}{120}&31&\multicolumn{2}{c|}{5810}&1623\\ 
		MELD&1038&114&280&9989&1109&2610\\
		DailyDialog&11118&1000&1000&87170&8069&7740\\
		EmoryNLP&713&99&85&9934&1344&1328\\
		\bottomrule
	\end{tabular}
	}
	\caption{The statistics of four datasets.}
	\label{tab:statistic}
\end{table}

We evaluate DAG-ERC on four ERC datasets. The statistics of them are shown in Table \ref{tab:statistic}.

\noindent\textbf{IEMOCAP} \citep{busso2008iemocap}: A multimodal ERC dataset.~Each conversation in IEMOCAP comes from the performance based on script by two actors. Models are evaluated on the samples with 6 types of emotion, namely \textit{neutral}, \textit{happiness}, \textit{sadness}, \textit{anger}, \textit{frustrated}, and \textit{excited}. Since this dataset has no validation set, we follow \citet{shen2020dialogxl} to use the last 20 dialogues in the training set for validation.

\noindent\textbf{MELD} \citep{poria2019meld}: A multimodal ERC dataset collected from the TV show \textit{Friends}. There are 7 emotion labels including \textit{neutral}, \textit{happiness}, \textit{surprise}, \textit{sadness}, \textit{anger}, \textit{disgust}, and \textit{fear}. 

\noindent\textbf{DailyDialog} \citep{li2017dailydialog}: Human-written dialogs collected from communications of English learners.~7 emotion labels are included: \textit{neutral},   \textit{happiness}, \textit{surprise}, \textit{sadness}, \textit{anger}, \textit{disgust}, and \textit{fear}. Since it has no speaker information, we consider utterance turns as speaker turns by default.

\noindent\textbf{EmoryNLP} \citep{zahiri2017emotion}:~TV show scripts collected from \textit{Friends}, but varies from MELD in the choice of scenes and emotion labels. The emotion labels of this dataset include \textit{neutral}, \textit{sad}, \textit{mad}, \textit{scared}, \textit{powerful}, \textit{peaceful}, and \textit{joyful}.

We utilize only the textual modality of the above datasets for the experiments. For evaluation metrics, we follow \citet{ishiwatari2020relation} and \citet{shen2020dialogxl} and choose micro-averaged F1 excluding the majority class (neutral) for DailyDialog and weighted-average F1 for the other datasets.

\subsection{Compared Methods}
We compared our model with the following baselines in our experiments:

\noindent \textbf{Recurrence-based methods:} DialogueRNN \citep{majumder2019dialoguernn}, DialogRNN-RoBERTa \citep{ghosal2020cosmic}, and COSMIC without external knowledge\footnote{In this paper, we compare our DAG-ERC with COSMIC without external knowledge, rather than the complete COSMIC, in order to make a clearer comparison on the model architecture, even though our DAG-ERC outperforms the complete COSMIC on IEMOCAP, DailyDialog and EmoryNLP.} \citep{ghosal2020cosmic}.

\noindent\textbf{Graph-based~methods:}~DialogurGCN \citep{ghosal2019dialoguegcn},~KET \citep{zhong2019knowledge}, DialogXL \citep{shen2020dialogxl} and RGAT \citep{ishiwatari2020relation}.

\noindent \textbf{Feature extractor:} RoBERTa \citep{liu2019roberta}. 

\noindent \textbf{Previous models with our extracted features:}  DialogueGCN-RoBERTa, RGAT-RoBERTa and DAGNN \citep{thost2021directed}\footnote{DAGNN is not originally designed for ERC, so we apply our DAG building method and the extracted feature for it.}.

\noindent \textbf{Ours:} DAG-ERC.

\section{Results and Analysis}

\subsection{Overall Performance}
The overall results of all the compared methods on the four datasets are reported in Table \ref{tab:overall}. We can note from the results that our proposed DAG-ERC achieves competitive performances across the four datasets and reaches a new state of the art on the IEMOCAP, DailyDialog and EmoryNLP datasets.

As shown in the table, when the feature extracting method is the same, graph-based models generally outperform recurrence-based models on IEMOCAP, DailyDialog, and EmoryNLP. 
This phenomenon indicates that recurrence-based models cannot encode the context as effectively as graph-based models, especially for the more important local context.~What's more, we see a significant improvement of DAG-ERC over the graph-based models on IEMOCAP, which demonstrates DAG-ERC's superior ability to capture remote information given that the dialogs in IEMOCAP are much longer (almost 70 utterances per dialog).

On MELD, however, we observe that neither graph-based models nor our DAG-ERC outperforms the recurrence-based models.~After going through the data, we find that due to the data collection method (collected from TV shows), sometimes two consecutive utterances in MELD are not coherent. Under this circumstance, graph-based models' advantage in encoding context is not that important. 

Besides, the graph-based models see considerable improvements when implemented with the powerful feature extractor RoBERTa.
In spite of this, our DAG-ERC consistently outperforms these improved graph-based models and DAGNN, confirming the superiority of the DAG structure and the effectiveness of the improvements we make to build DAG-ERC upon DAGNN.

\begin{table}[t]
	\centering
		\resizebox{0.48\textwidth}{!}{
	\begin{tabular}{l|cccc} 
		\toprule
		Model & IEMOCAP &MELD &DailyDialog &EmoryNLP\\ 
		\hline
	   DialogueRNN &   62.75   &57.03  &-           &-\\
	   \ \ +RoBERTa & 64.76 & 63.61 & 57.32          &37.44\\
		COSMIC &63.05 	& \textbf{64.28}  & 56.16  &37.10\\
		\hline
		KET&59.56 &58.18 & 53.37           & 33.95 \\
		DialogXL &65.94 &62.41 &54.93           &34.73\\
	    DialogueGCN &64.18 & 58.10 &-           &-\\
		\ \ +RoBERTa&64.91 &63.02 &57.52           &38.10\\
		RGAT &65.22 &60.91 &54.31          &34.42\\
		\ \ +RoBERTa&66.36 &62.80 &59.02          &37.89\\
		\hline
		RoBERTa &63.38 &62.88 &58.08           &37.78\\
		\hline
		DAGNN &64.61 &63.12 &58.36           &37.89\\
	     \rowcolor{mygray}DAG-ERC &\textbf{68.03} &63.65 &\textbf{59.33}           &\textbf{39.02}\\
		\bottomrule
	\end{tabular}
	}
	\caption{Overall performance on the four datasets.}
	\label{tab:overall}
\end{table}

\subsection{Variants of DAG Structure}\label{sec:variants}

In this section, we investigate how the structure of DAG would affect our DAG-ERC's performance by applying different DAG structures to DAG-ERC. 
In addition to our proposed structure, we further define three kinds of DAG structure: 
(1) sequence, in which utterances are connected one by one; 
(2) DAG with single local information, in which each utterance only receives local information from its nearest neighbor, and the remote information remains the same as our DAG;
(3) common DAG, in which each utterance is connected with $\kappa$ previous utterances.  
Note that if there are only two speakers taking turns to speak in a dialog, then our DAG is equivalent to common DAG with $\kappa = 2\omega$, making the comparison less meaningful.~Therefore, we conduct the experiment on EmoryNLP, where there are usually multiple speakers in one dialog, and the speakers speak in arbitrary order. The test performances are reported in Table \ref{tab:variants}, together with the average number of each utterance's predecessors.

Several instructive observations can be made from the experimental results.~Firstly, the performance of DAG-ERC drops significantly when equipped with the sequence structure. 
Secondly, our proposed DAG structure has the highest performance among the DAG structures. 
Considering our DAG with $\omega=2$ and common DAG with $\kappa=6$, with very close numbers of predecessors, our DAG still outperforms the common DAG by a certain margin.
This indicates that the constraints based on speaker identity and positional relation are effective inductive biases, and the structure of our DAG is more suitable for the ERC task than rigidly connecting each utterance with a fixed number of predecessors.
Finally, we find that increasing the value of $\omega$ may not contribute to the performance of our DAG, and $\omega=1$ tends to be enough.

\begin{table}[t]
	\centering
	\small
	\resizebox{0.48\textwidth}{!}{
	\begin{tabular}{l|p{1.5cm}<{\centering}p{1.5cm}<{\centering}}
		\toprule
		DAG  & \# Preds & F1 score\\
		\hline
		Sequence & 0.92 & 37.57 \\
		Single local information  &1.66 & 38.22\\
		\hline
		Common $\kappa=2$ &1.78 &  38.30\\
		Common $\kappa=4$ &3.28 &  38.34\\
		Common $\kappa=6$ &4.50 &  38.48\\
		\hline
		Ours $\omega=1$ &2.69 &  \textbf{39.02}\\
		Ours $\omega=2$ &4.46 &  38.90\\
		Ours $\omega=3$ &5.65 &  38.94\\
		\bottomrule
	\end{tabular}
	}
	\caption{Different DAGs applied to DAG-ERC.}
	\label{tab:variants}
\end{table}

\begin{table}[b]
	\centering
	\resizebox{0.48\textwidth}{!}{
	\begin{tabular}{l|cccc}
		\toprule
		Method  &IEMOCAP & MELD &DailyDialog &EmoryNLP\\
		\hline
		DAG-ERC &68.03& 63.65& 59.33& 39.02\\
		w/o rel-trans&64.12 ($\downarrow$3.91)& 63.29 ($\downarrow$0.36) & 57.12 ($\downarrow$2.21) & 38.87 ($\downarrow$0.15) \\ 
		w/o $\widetilde{H}$& 66.19 ($\downarrow$1.84) &  63.17 ($\downarrow$0.48)  & 58.05 ($\downarrow$1.28) & 38.54 ($\downarrow$0.48) \\
		w/o $C$ & 66.32 ($\downarrow$1.71) &  63.36 ($\downarrow$0.29) & 58.90 ($\downarrow$0.43)& 38.50 ($\downarrow$0.52) \\ 
		\bottomrule
	\end{tabular}
	}
	\caption{Results of ablation study on the four datasets, with \emph{rel-trans}, $\widetilde{H}$, and $C$ denoting relation-aware feature transformation, nodal information unit, and contextual information unit, respectively.}
	\label{tab:ablation}
\end{table}

\subsection{Ablation Study}
 To study the impact of the modules in DAG-ERC, we evaluate DAG-ERC by removing relation-aware feature transformation, the nodal information unit, and the contextual information unit individually. 
The results are shown in Table \ref{tab:ablation}. 

As shown in the table, removing the relation-aware feature transformation causes a sharp performance drop on IEMOCAP and DailyDialog, while a slight drop on MELD and EmoryNLP.
Note that there are only two speakers per dialog in IEMOCAP and DailyDialog, and there are usually more than two speakers in dialogs of MELD and EmoryNLP. 
Therefore, we can infer that the relation of whether two utterances have the same speaker is sufficient for two-speaker dialogs, while falls short in the multi-speaker setting.

Moreover, we find that on each dataset, the performance drop caused by ablating nodal information unit is similar to contextual information unit, and all these drops are not that critical.
This implies that either the nodal information unit or contextual information unit is effective for the ERC task, while combining the two of them can yield further performance improvement.

\begin{figure}[t]
	\centering
	\includegraphics[scale=0.53]{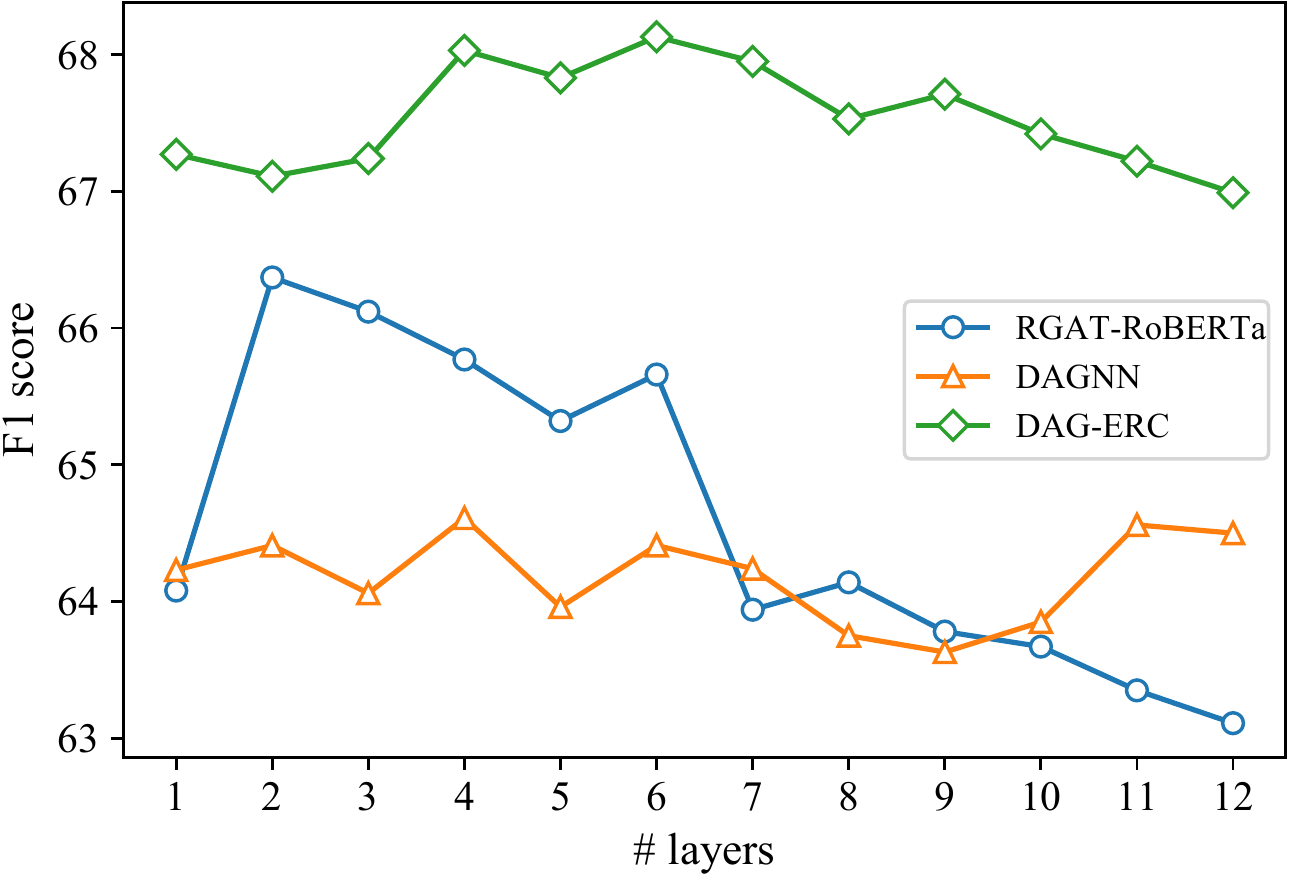} 
	\caption{Test results of RGAT-RoBERTa, DAGNN, and DAG-ERC on the IEMOCAP dataset by different numbers of network layers.}
	\label{fig:Layers}
\end{figure} 

\subsection{Number of DAG-ERC Layers} 
According to the model structure introduced in Section \ref{sec:GNN_RNN_DAGNN}, the only way for GNNs to receive information from a remote utterance is to stack many GNN layers. However, it is well known that stacking too many GNN layers might cause performance degradation due to over-smoothing \citep{kipf2016semi}.~We investigate whether the same phenomenon would happen when stacking many DAG-ERC layers.~We conduct an experiment on IEMOCAP and plot the test result by different numbers of layers in Figure \ref{fig:Layers}, with RGAT-RoBERTa and DAGNN as baselines. 
As illustrated in the figure, RGAT suffers a significant performance degradation after the number of layers exceeds 6. While for DAGNN and DAG-ERC, with the number of layers changes, both of their performances fluctuate in a relatively narrow range, indicating that over-smoothing tends not to happen in the directed acyclic graph networks.

\subsection{Error Study}
After going through the prediction results on the four datasets, we find that our DAG-ERC fails to distinguish between similar emotions very well, such as \textit{frustrated} vs \textit{anger}, \textit{happiness} vs \textit{excited}, \textit{scared} vs \textit{mad}, and \textit{joyful} vs \textit{peaceful}. 
This kind of mistake is also reported by \citet{ghosal2019dialoguegcn}. 
Besides, we find that DAG-ERC tends to misclassify samples of other emotions to \textit{neutral} on MELD, DailyDialog and EmoryNLP due to the majority proportion of \textit{neutral} samples in these datasets.
\begin{table}[t]
	\centering
	\resizebox{0.5\textwidth}{!}{
	\begin{tabular}{l|c|c|c|c}
		\toprule
		\multirow{2}*{Dataset} & \multicolumn{2}{c|}{Emotional shift} &\multicolumn{2}{c}{w/o Emotional shift}\\ 
		&\# Samples& Accuracy &\# Samples&Accuracy\\
		\hline
		IEMOCAP&576 &57.98\% &1002 &74.25\%\\ 
		MELD&1003 &59.02\% &861 &69.45\%\\
		DailyDialog&670 &57.26\% &454 &59.25\%\\
		EmoryNLP&673 &37.29\% &361 &42.10\%\\
		\bottomrule
	\end{tabular}
	}
	\caption{Test accuracy of DAG-ERC on samples with emotional shift and without it.}
	\label{tab:emotion_shift}
\end{table}

We also look closely into the \emph{emotional shift} issue, which means the emotions of two consecutive utterances from the same speaker are different. 
Existing ERC models generally work poorly in emotional shift.~As shown in Table \ref{tab:emotion_shift}, our DAG-ERC also fails to perform better on the samples with emotional shift than that without it, though the performance is still better than previous models. For example, the accuracy of DAG-ERC in the case of emotional shift is 57.98\% on the IEMOCAP dataset, which is higher than 52.5\% achieved by DialogueRNN \citep{majumder2019dialoguernn} and 55\% achieved by DialogXL \citep{shen2020dialogxl}.

\section{Conclusion}
In this paper, we presented a new idea of modeling conversation context with a directed acyclic graph (DAG) and proposed a directed acyclic graph neural network, namely DAG-ERC, for emotion recognition in conversation (ERC). Extensive experiments were conducted and the results show that the proposed DAG-ERC achieves comparable performance with the baselines. Moreover, by comprehensive evaluations and ablation study, we confirmed the superiority of our DAG-ERC and the impact of its modules.~Several conclusions can be drawn from the empirical results.~First, the DAG structures built from conversations do affect the performance of DAG-ERC, and with the constraints on speaker identity and positional relation, the proposed DAG structure outperforms its variants.~Second, the widely utilized graph relation type of whether two utterances have the same speaker is insufficient for multi-speaker conversations. Third, the directed acyclic graph network does not suffer over-smoothing as easily as GNNs when the number of layers increases. Finally, many of the errors misjudged by DAG-ERC can be accounted for by similar emotions, neutral samples and emotional shift. These reasons have been partly mentioned in previous works but have yet to be solved, which are worth further investigation in future work.

\section*{Acknowledgments}
We thank the anonymous reviewers.~This paper was supported by the Program for Guangdong Introducing Innovative and Entrepreneurial Teams (No.2017ZT07X355).

\bibliographystyle{acl_natbib}
\bibliography{acl2021}


\end{document}